\documentclass[sn-apa]{sn-jnl}

\usepackage{graphicx}%
\usepackage{todonotes}%
\usepackage{multirow}%
\usepackage{amsmath,amssymb,amsfonts}%
\usepackage{amsthm}%
\usepackage{mathrsfs}%
\usepackage[title]{appendix}%
\usepackage{xcolor}%
\usepackage{textcomp}%
\usepackage{manyfoot}%
\usepackage{booktabs}%
\usepackage{algorithm}%
\usepackage{algorithmicx}%
\usepackage{algpseudocode}%
\usepackage{listings}%
\usepackage{natbib}
\usepackage{array}
\usepackage{booktabs}
\usepackage{tabularx}
\usepackage{xcolor}
\usepackage{colortbl}
\usepackage{siunitx}
\usepackage{float}%
\usepackage{graphicx}
\usepackage{subcaption}  

\begin{document}

\title[Article Title]{LLM-based feature generation  from text for interpretable machine learning}

\author[1]{\fnm{Vojtěch} \sur{Balek}}\email{balv05@vse.cz}

\author[1]{\fnm{Lukáš} \sur{Sýkora}}\email{lukas.sykora@vse.cz}

\author[1]{\fnm{Vilém} \sur{Sklenák}}\email{vilem.sklenak@vse.cz}

\author*[1]{\fnm{Tomáš} \sur{Kliegr}}\email{tomas.kliegr@vse.cz}

\affil*[1]{\orgdiv{Department of Information and Knowledge Engineering},\orgname{Prague University of Economics and Business}, \orgaddress{\street{nam W Churchilla 
 4}, \city{Prague}, \postcode{13067}, \country{Czech Republic}}}


\abstract{
Traditional text representations like embeddings and bag-of-words hinder rule learning and other interpretable machine learning methods due to high dimensionality and poor comprehensibility. This article investigates using Large Language Models (LLMs) to extract a small number of interpretable text features. We propose two workflows: one fully automated by the LLM (feature proposal and value calculation), and another where users define features and the LLM calculates values. This LLM-based feature extraction enables interpretable rule learning, overcoming issues like spurious interpretability seen with bag-of-words. 
We evaluated the proposed methods on five diverse datasets (including scientometrics, banking, hate speech, and food hazard). LLM-generated features yielded predictive performance similar to the SciBERT embedding model but used far fewer, interpretable features. Most generated features were considered relevant for the corresponding prediction tasks by human users. We illustrate practical utility on a case study focused on mining recommendation action rules for the improvement of research article quality and citation impact.
}




\maketitle

\section{Introduction}\label{sec1}
Traditional text representations, such as bag of words (BoW) and embeddings, have posed significant challenges for “white-box” machine learning models. Using rule learning as an example, the use of BoW representation results in many features tied to specific words, leading to rules that are difficult to interpret and over-specific. On the other hand, embeddings make it practically impossible to derive any rules due to their complex, dense representations. 
The use of text in machine learning thus typically requires an explanation algorithm, while direct learning of an interpretable model is preferred \citep{atzmueller2024explainable}. 

Large language models (LLMs) have seen a significant increase in use in academic and commercial spheres alike, and the wide availability of open-weight models such as  Llama2  \citep{touvron2023llama} makes LLMs suitable for use as a component in machine learning workflows. These models are commonly used for text summarization, classification, or natural language generation.

This article explores whether large language models (LLMs) could address this by extracting a small number of interpretable features from text. We propose two workflows, one fully automated, where LLM both proposes suitable features and computes their values, and a second one where the feature names are user-specified and LLM only generates their values. We further show that LLM-based feature extraction can be used as an input for rule learning and has well-interpretable results. We evaluate this process on two datasets (CORD-19 and M17+) from the scientometric domain and three additional datasets from other domains: customer intent from the banking domain, hate speech and food hazard.

To illustrate the process, consider the scholarly document quality prediction use case. First, we use an LLM to assess a large number of scientific article abstracts, evaluating them according to multiple criteria.  These criteria can either be LLM-generated or user-specified, as in the example below.

\vspace{2mm}
\noindent\fbox{%
    \parbox{\textwidth}{%
\emph{Example.} Our LLM feature generator applied to a hypothetical initial version of this articles' abstract generated features such as  \emph{rigor=high} (on a three point scale low/medium/high), \emph{grammar=0} (on binary scale, where 0 is free of errors and 1 is contains errors), \emph{novelty=low}, \emph{accessibility=high}, \emph{replicability=0} (not good). Action rules were learnt on a training dataset with similarly LLM-annotated features, and a target corresponding to human expert assessment of article quality. 
An applicable action rule found for our abstract is:

\vspace{-8mm}
\begin{multline*}
  r_{15} : \text{novelty}= (\text{low} \rightarrow \text{high} ) \wedge \text{replicability}= (0 \rightarrow 1 ) \Rightarrow \text{evaluation}= (\text{avg} \rightarrow \text{best} ).
\end{multline*}
\vspace{-6mm}

This rule, which used only two of the available features, indicates that in its current form, the article is predicted to receive an average expert evaluation, but if replicability and novelty are improved, the probability of the evaluation increasing to a higher rating will rise.\footnote{The full listing of  $r_{15}$ in  Section~\ref{sec:action:results}) also includes uplift, an indicator of probabilistic nature of this recommendation.
} Consequently, we could modify the abstract to emphasize that LLM-based features have not yet been used in rule learning and that our use of an open LLM fosters replicability compared to prior work that used closed commercial LLMs.

}}

The first contribution of our work is in a multi-pronged evaluation of the quality of LLM-generated features, including their agreement with domain expertise, analysis of interpretability, and predictive performance. As reference methods, we use both state-of-the-art but  ``black-box'' embedding-based methods as well as the partly interpretable BoW representation. We also demonstrate the effect of the fusion of LLM-based and BoW features. 

In addition to the quantitative results, as the second contribution,  we demonstrate how LLM-based features can be used to build actionable rule-based predictions using action rules \citep{ras_action-rules_2000}. We chose this specific recommendation task over the more commonly used classification task because it is one area where rule learning can provide a distinct functionality.
Action rule-learning is a method for deriving actionable insights from data by identifying specific changes that can lead to desired outcomes. In this work, action rules are used as counterfactual (what-if) explanations to gain a deeper understanding of the factors that contribute to successful article evaluations or citation rates. By identifying specific changes in attributes that could improve these outcomes, action rules provide valuable insights into the underlying reasons for success. Considering our domain of research article impact,  this can lead to recommendations for the article writing process, resulting in better evaluations or higher citation rates.

This article is organized as follows. Section~\ref{sec:methods} presents the methodology, including the two datasets used and the description of LLM-based feature generation methodology and methods for feature quality validation. This part also introduces the concept of action rule learning. Section~\ref{sec:results} presents the results. This proceeds from the description of the generated datasets to statistical tests for feature significance, analysis of their predictive performance and SHAP-value analysis of built classifiers. The last part of this section covers the results of action rule mining. Discussion and limitations are present in Section~\ref {sec:discussion} and related work in Section~\ref{sec:relatedwork}. The conclusions summarize the main results and suggest possible future extensions. 

\section{Methods}\label{sec:methods}
In this research, we employ the publicly available open-source Llama2 model and use it to extract several features from scientific article abstracts. These features are then used as inputs for several machine-learning algorithms, and their predictive performance is measured against alternative forms of text representations, such as word embeddings or TF-IDF matrix.

\subsection{Input data}
Five datasets were used for experiments in this work. Two datasets (CORD-19 and M17+) contain scientific article titles, abstracts, and a target variable related to research impact. 
To demonstrate the universality of the research, three additional datasets from fields other than scientometrics are utilized: BANKING77\footnote{\url{https://paperswithcode.com/dataset/banking77}}, Hate Speech\footnote{\url{https://paperswithcode.com/dataset/hate-speech}} and Food Hazard\footnote{\url{https://food-hazard-detection-semeval-2025.github.io/}}.

\paragraph{Scientometric datasets}
\emph{The CORD-19 dataset} is the first dataset of interest. It is primarily composed of scientific articles related to coronaviruses that span from the 1970s to 2020, the beginning of the COVID-19 outbreak. The target variable in this dataset represents either a low or high citation rate. This variable was computed in previous research by \citet{Beranova2022} by dividing the number of citations obtained from OpenCitations and Web of Science by the age of the publication in years and then assigning 0 to articles below the median and 1 to articles above the median. In order to reduce the computational complexity, the dataset size was reduced to 3,000 articles and was balanced with respect to the target variable. 

From both datasets, we use article abstracts as the main source of information.  The use of abstracts improves the applicability of our method, as abstracts are more easily available.  

\emph{The M17+ dataset} \citep{Cap2024thesis} is also used in this work's experiments. The dataset consists of 7,710 articles from Czech research institutions that were graded by a jury of experts in a program subsidized by the government of the Czech Republic, similar to the U.K. Research Excellence Framework 2021 (REF) that has been the subject of previous research with regards to grade prediction of articles by machine-learning methods \citep{refpred}. The target variable in this dataset is an ordinal variable ranging from 1 to 5, with 1 meaning that the article is of world-class quality and impact, while 5 means that the article is of mediocre quality and impact. The dataset was downsampled from 7,710 articles to 2,000 with respect to the target variable. This resulted in an even distribution of 400 articles belonging to each class. 

Both scientometric datasets contained several columns beside the article title, abstract, and target variable, such as journal name, impact factors (IF), or article influence score (AIS). The final version of both datasets contained only 3 columns - title, abstract, and target. 
\paragraph{Other datasets}  
\emph{The BANKING77 dataset}\footnote{\url{https://paperswithcode.com/dataset/banking77}} provides a fine-grained set of intents in the banking domain. It comprises 13,083 customer service queries labeled with 77 intents. This dataset focuses on fine-grained single-domain intent detection and is commonly used for intent classification tasks \citep{casanueva2020efficient}.

\emph{The Hate Speech dataset}\footnote{\url{https://paperswithcode.com/dataset/hate-speech}} consists of sentences extracted from Stormfront, a white supremacist forum. A total of 10,568 sentences have been manually annotated as conveying hate speech or not. This dataset is utilized for research in hate speech detection and related tasks \citep{de2018hate}.

\emph{The Food Hazard dataset}\footnote{\url{https://food-hazard-detection-semeval-2025.github.io/}} is a part of the Semeval-2025 challenge and contains 6,644 food incident reports containing information about product recalls from regulatory bodies along with 4 target variables that correspond to 4 classification tasks. We chose the task of predicting the 'hazard-category' target variable.

All datasets were preprocessed to retain only the relevant textual data and target variables necessary for our experiments.

\subsection{Machine-learning algorithms and their hyperparameters}\label{sec:algos}
For the purposes of a thorough comparison of predictive performance, we utilized various machine-learning algorithms for each feature subset.

The Scikit-learn Python library was used as a source for a range of models \citep{scikit-learn}. Specifically, we tested the predictive performance of the Random Forest Classifier, Gradient Boost Classifier, Ada Boost Classifier, Support Vector Machines, and Logistic Regression on CORD-19 and M17+ datasets. For the ensemble models, the best models were chosen from the hyperparameter space in Table \ref{tab:hyperparameter_space} using grid search. Out of the aforementioned models, the Gradient Boost Classifier consistently showed the best results. Therefore, we used only this model for subsequent experiments with LLM and BoW features on M17+ and CORD-19 datasets. For experiments on BANKING-77, Hate Speech, and Food Hazard datasets, which were several times larger than M17+ and CORD-19, we used the Random Forest Classifier, which resulted in a very similar performance to the Gradient Boost Classifier but was faster.

Settings for TF-IDF, which served as one of the baselines, are in the provided GitHub repository.
\begin{table}[!ht]
    \centering
    \begin{tabular}{>{\raggedright\arraybackslash}p{4cm} >{\raggedright\arraybackslash}p{4cm} >{\raggedright\arraybackslash}p{4cm}}
        \toprule
        \textbf{Hyperparameter} & \textbf{GradientBoost Values} & \textbf{RandomForest Values} \\
        \midrule
        Max Depth & 10, 20, 30, 40, 45, 50 & 3, 5, 10, None \\
        \addlinespace
        Max Features & log2, sqrt, 30 & -- \\
        \addlinespace
        Min Samples Leaf & 30, 40, 50, 60, 70, 90 & 1, 2, 5 \\
        \addlinespace
        Min Samples Split & 20, 30, 50, 60, 90 & 2, 5, 10 \\
        \addlinespace
        Number of Estimators & 100, 200, 400, 1000 & 50, 100, 200 \\
        \bottomrule
    \end{tabular}
    \caption{Hyperparameter space for ensemble models}
    \label{tab:hyperparameter_space}
\end{table}

In addition to scikit-learn, for experiments on the two scientometric datasets we used the 
\textit{AutoGluon} library. \textit{AutoGluon} is a deep-learning API used for training neural networks on a combination of tabular and image, text, or audio data \citep{tang2024autogluon}. In this study, we employed the AutoGluon Tabular model with title, abstract and LLM-generated features as input. We tested both regressor and classifier configurations. The classifier configuration provided better results in all observed metrics and was thus used as the only variation of the AutoGluon model.

\subsection{LLM-based feature generation based on user-set feature list}\label{sec:features:derivation}
The first method of LLM-assisted feature generation is based on one-shot generation with user-constructed prompts for manually selected features. This method was used for the extraction of features for CORD-19 and M17+ datasets only. The features shown in Table \ref{tab:abstract_criteria} were manually selected in this experiment by the authors. The selection of feature names was based on our knowledge of the subject domain from previous publications \citep{Beranova2022,dvorackova2024explaining} that also focused on inspecting the relationship between article content and research impact.

\begin{table}[h!]
    \centering
    \begin{tabularx}{\textwidth}{>{\raggedright\arraybackslash}p{3cm} X >{\raggedright\arraybackslash}p{4cm}}
        \toprule
        \textbf{Criteria} & \textbf{Description} & \textbf{Values} \\
        \midrule
        Rigor & Assessed methodological soundness of logic presented in the abstract & \{low, medium, high\} \\
        \addlinespace
        Novelty & Assessed innovativeness of the research based on the abstract & \{low, medium, high\} \\
        \addlinespace
        Accessibility & Assessed understandability of the language used in the abstract & \{low, medium, high\} \\
        \addlinespace
        Replicability & Assessed the authors' mention of the reproducibility of results presented in the abstract & \{no, yes\} \\
        \addlinespace
        Research type & Assessed type of research types and methods mentioned in the abstract & 16 research types \\
        \addlinespace
        Discipline & Assessed all disciplines that the article concerns & 41 FORD general categories \citep{/content/publication/9789264239012-en} \\
        \addlinespace
        Grammar & Assessed the presence of grammar errors in the abstract & \{no, yes\} \\
        \bottomrule
    \end{tabularx}
    \caption{User-specified features for scientometric datasets. These features were used for LLM-based generation of a total of 62 features (3 ordinal, 59 binary) from article abstracts in M17+ and CORD-19 datasets.}
    \label{tab:abstract_criteria}
\end{table}

The user-defined features included several ordinal features, such as rigor.
The \textit{research type} and \textit{discipline} features were originally multinomial and were divided into multiple binary variables indicating the articles belonging to respective research types or disciplines. Each article was allowed to belong to more than one discipline and research.

Listing \ref{ex:llm-prompt} shows an example of a user-constructed prompt used for LLM-based generation of the value of a specific feature (here 'rigor') for a given data instance (a research paper abstract).
\begin{lstlisting}[float,floatplacement=H,caption={Example prompt for LLM-based determination of the value of a specific feature}, label={ex:llm-prompt}, breaklines=true, 
  basicstyle=\ttfamily, 
  columns=flexible, 
  frame=single, 
  xleftmargin=20pt, 
  linewidth=\textwidth]
You are a categorization assistant. Your job will be to assign a certain characteristic to a research paper based on its abstract.  
    
        In this instance you will assess the methodological rigor of the research. 
    
        Methodological rigor represents the soundness of logic presented in the article abstract by its authors. 
    
        You will choose between three levels of rigor: low, medium and high. Low being the least rigorous and high being the most.  
    
        Be concise, no explanation is to be provided. Your answer will consist of an answer in plain json format and nothing else like so:
    
        {{ 
    
            ""rigor"": ""value"" 
    
        }} 
    
        Abstract to be evaluated:
        <abstract>
\end{lstlisting}

We used the  Llama2 13B GPTQ Chat\footnote{https://huggingface.co/TheBloke/Llama-2-13B-chat-GPTQ} model through the transformers library API \citep{DBLP:journals/corr/abs-1910-03771}, which allowed us to set model hyperparameters so that the models produced consistent results. The hyperparameter \texttt{do\_sample} in the \texttt{model.generate()} method was set to False, as that ensured deterministic generation by always selecting the most likely next token, avoiding randomness in the output.

\subsection{LLM-based feature generation with automatic feature discovery}
\label{sec:automatic}

In contrast to the previous method, which used LLMs solely for the generation of feature values, this section introduces an enhanced automated workflow leveraging LLMs for feature discovery --  determining the names of features to extract.
This method further reduces the need for human intervention and was applied to all five datasets.

The core of the approach is a new prompt that contains essential details about the dataset, including its name, a concise description, the text column name, the target column name, and a definition of the target variable. The full prompt is depicted in Figure~\ref{fig:fullprompt} (Appendix~\ref{secA1}).
As a variable part of the prompt, the model receives a representative sample of rows from the dataset. Based on this information, the model autonomously discovers relevant features for extraction along with corresponding prompts designed explicitly to facilitate their extraction.

The subsequent phase for generating the value of a specific feature for a given data instance is done again automatically with an LLM prompt. The example of a feature generation prompt is depicted in Figure~\ref{fig:genprompt} (Appendix~\ref{secA1}). The difference from the approach described in the previous Section \ref{sec:features:derivation} and exemplified in Listing~\ref{ex:llm-prompt} is that now not only is the feature value generated (determined, computed) automatically but also the prompt used to generate the feature value is created automatically for a given feature type. 

For comparison with the method introduced in the previous subsection, examples of both levels of the automation of feature generation are included in Tables~\ref{tab:c19_table_sample}-\ref{tab:bank77_table_sample} in Section~\ref{ss:llm-augmented-datasets} covering LLM-augmented datasets.
Additionally, a direct comparison in terms of predictive performance (SHAP values) is included in the Results section in Figure~\ref{fig:shap-manual-vs-auto-m17}.

In our experiments, automated feature discovery was performed using the model \texttt{gpt-4o-2024-08-06} (temperature: 0, top\_p: 0.9). GPT-4o, introduced by OpenAI in May 2024, is a multimodal large language model capable of processing and generating text, images, and audio.  We used 40 randomly selected rows as part of the customization of the feature selection prompt.  Following the discovery phase, feature generation was executed using the OpenAI Batch API with the model \texttt{gpt-4o-mini-2024-07-18} (temperature: 0, top\_p: 0.9). To optimize cost efficiency, a 24-hour processing window was chosen.

\subsection{Explainability}
In this study, we used the SHAP (SHapley Additive exPlanations) Python package  \citep{NIPS2017_7062} to enhance the interpretability of features generated by the LLM. SHAP uses Shapley values from game theory to explain the contribution of each feature to the model's predictions. This allowed us to identify the most important LLM-generated features and understand their individual impact on the model's performance. By integrating SHAP, we made our machine learning models more transparent and easier to interpret, ensuring the effectiveness and relevance of the LLM-generated features.

\subsection{Feature subsets}
To compare predictive performance in the classification task, we generated various feature subsets. 

For the sake of reproducibility and comparability, CORD-19, M17+ and Hate Speech sets were split into train and test sets in an 80:20 ratio with a fixed random seed before the application of feature subsetting, while BANKING77 and Food Hazard were already split from the dataset source. This resulted in the same respective instances being tested each time, only with different features.

For all datasets, we experimented with the following feature set versions:
\begin{itemize}
    \item \textit{BoW only} is a feature set where the only features used are bag-of-words terms weighted by their TF-IDF score along with the target variable. 
    \item \textit{LLM-features only} dataset has all columns beside the target variable LLM-generated. 
    \item \textit{BoW + LLM-generated features} is a dataset with TF-IDF-weighted terms and all LLM-features appended.
\end{itemize}

Additionally, for two datasets from the scientific  domain (CORD-19, M17+), we used the following three additional feature sets for experiments focused on user-specified features:
\begin{itemize}
\item \textit{text only} datasets  we used only \textit{title} and \textit{abstract} as input in the AutoGluon Tabular model.  
\item \textit{SciBERT embeddings only} is a feature set where each abstract was embedded using the SciBERT model  \citep{beltagy2019scibert}.  SciBERT was trained on a large amount of scientific data and significantly outperformed the base BERT model \citep{devlin-etal-2019-bert}.  The embeddings produced by this model
are vectors of length 768.
\item \textit{text + LLM-generated features} combines \textit{title} and \textit{abstract} text columns from the original datasets with LLM-generated features. This feature set is used with the AutoGluon Tabular model. 
\end{itemize}

Detailed settings of all baseline methods can be found on the GitHub page referenced from the Availability of code and data section.

\subsection{Action rules}
\label{sec:actionrules}
An action rule is typically generated by combining two classification rules \citep{ras_action-rules_2000}, each indicating different outcomes. The classification rules, which serve as the basis for action rules, can be extracted using algorithms like Apriori \citep{agrawal1994fast} modified to suit classification tasks. A formal representation of the classification rule is in Equation \ref{eq:formal}. 

\vspace{-4mm}
\begin{equation}
r_i: \phi \Rightarrow \psi, 
\label{eq:formal}
\end{equation}
\vspace{-4mm}

where the antecedent $\phi$ is a set containing at least one item and the consequent $\psi$ contains one item. A set of items is also referred to as an \emph{itemset}. The absolute support of a classification rule corresponds to the number of transactions (here articles) $t$ in a dataset $D$ containing all items (conditions) from the antecedent $\phi$  and the consequent $\psi$ (target) of the given rule, see Equation \ref{eq:support}.  

\vspace{-4mm}
\begin{equation}
    sup(\phi \Rightarrow \psi) = |t \in D: \phi \subset  t \land \psi \subset t   |.
\label{eq:support}    
\end{equation} 
\vspace{-4mm}

The confidence of a rule corresponds to the ratio between the number of articles $t$ that contain all items from both the antecedent $\phi$ and the  consequent $\psi$ to the number of articles $t$ that contain all items from the antecedent $\phi$, see Equation \ref{eq:confidence}.

\vspace{-2mm}
\begin{equation}
    \textit{conf}(\phi \Rightarrow \psi) = \frac{sup(\phi \Rightarrow \psi)}{|t \in D: \phi \subset t|}
\label{eq:confidence}    
\end{equation} 
\vspace{-2mm}

For example, the following classification rules, denoted as \(r_1\) (see Equation \ref{eq:r1}) and \(r_2\) (see Equation \ref{eq:r2}), can be extracted from the M17+ dataset (enriched with new features from Section \ref{sec:features:derivation}) if the minimum support is set to 40 and the minimum confidence is set to 0.7:

\vspace{-8mm}
\begin{multline}
r_1 : \text{area = chemistry} \wedge \text{rigor = medium} \Rightarrow \text{evaluation = bad}
\\ 
\textit{with\:support\:50\:and\:confidence\:71.4\%.} 
\label{eq:r1}
\end{multline}

\vspace{-12mm}

\begin{multline}
r_2 :  \text{area = chemistry}  \wedge  \text{rigor = high} \Rightarrow \text{evaluation = good}
\\ 
\textit{with\:support\:249\:and\:confidence\:71.3\%. }
\label{eq:r2}
\end{multline}
\vspace{-6mm}

From these classification rules \(r_1\) and \(r_2\), an action rule \(r_3\) can be constructed. In this case, the 'area' attribute is considered stable, meaning that it does not allow for changes (authors are not advised to change their research area). The 'rigor' attribute, however, is flexible, allowing for recommended changes that could lead to a reclassification of the object to a desired outcome.

\vspace{-8mm}
\begin{multline}
  r_3 : \text{area = chemistry} \wedge \text{rigor}= (\text{medium} \rightarrow \text{high} ) \Rightarrow \text{evaluation}= (\text{bad} \rightarrow \text{good} )
  \\ 
  \textit{with\:uplift\:15.0\%}. 
\label{eq:action_rule_r3}  
\end{multline}
\vspace{-6mm}

This action rule \(r_3\) (see Equation \ref{eq:action_rule_r3}) suggests that if all articles in the chemistry research area in the training data could improve their methodological rigor from medium to high, the overall probability of receiving a good evaluation (rated as 1 or 2) would increase by 15\%.  In practical terms, this could mean that approximately 300 articles would transition from poor evaluations (rated as 4 or 5) to good evaluations (rated as 1 or 2). 

An action rule \( r_{a} \)  can be decomposed into two classification rules, \( r_{undesired} \) and \( r_{desired} \): one that predicts the undesired state of the target (before intervention), and another that predicts the desired state (after intervention), represented as \( r_{undesired} \rightarrow r_{desired} \). Uplift is a measure used to predict the incremental response to an action \citep{radcliffe2007using}, see Equation \ref{eq:uplift}. Unlike commonly used measures such as confidence and lift, the uplift measure reflects the change over the entire dataset. Therefore, even rules with a relatively small value of uplift can be of practical significance.

\vspace{-5mm}
\begin{equation}
\begin{split}
\textit{uplift($r_a$)} & = P(\text{decision} \mid \text{treatment}) - P(\text{decision} \mid \text{no treatment}) \\
& = \frac{(\textit{conf}_{r_{desired}} - (1 - \textit{conf}_{r_{undesired}})) * \frac{supp_{r_{undesired}}}{\textit{conf}_{r_{undesired}}}}{|t \in D: t|}
\label{eq:uplift}  
\end{split}
\end{equation}
\vspace{-4mm}

The instances are divided into two groups: control and exposed. The exposed group (\textit{treatment}) is subjected to the recommended action, whereas the control group (\textit{no treatment}) does not receive the recommended action. The \textit{decision} represents the classification of the instances.

This example demonstrates how action rules can be used to derive counterfactual (what-if) explanations, providing insights into how specific changes in attributes might lead to better outcomes. Such rules can also be employed to recommend improvements for future research, guiding the refinement of methodologies to achieve better results. 

During the generation of action rules, it is necessary to address the issue of generating a large number of highly similar rules. The \textit{dominant action rule} approach to reducing action rules is inspired by \textit{closed association rules} \citep{pasquier1999discovering}. This concept is analogous to \textit{dominant action rules} in that closed rules cannot be extended without a loss of support, which can be likened to the reduction in uplift seen in dominant rules. A \textit{dominant action rule} is one that cannot be further expanded with additional conditions to increase its uplift. For example, the rule \(r_4\) (see Equation \ref{eq:dominant_rule_example}) is dominant over the longer rule \(r_5\). In this specific case, the conditions in rule \(r_4\) are not extended in \(r_5\) in a way that increases uplift. Instead, the addition of another condition in \(r_5\) results in a decreased uplift, demonstrating the dominance of \(r_4\) over \(r_5\).

For example, the rule \(r_4\) (see Equation \ref{eq:dominant_rule_example}) is a dominant rule for the rule \(r_5\) (see Equation \ref{eq:non_dominant_rule_example}).

\vspace{-8mm}
\begin{multline}
  r_4 : \text{rigor}= (\text{medium} \rightarrow \text{high} ) \Rightarrow \text{evaluation}= (\text{bad} \rightarrow \text{good} )
  \\ 
  \textit{with\:uplift\:16.96\%}. 
\label{eq:dominant_rule_example}  
\end{multline}

\vspace{-12mm}

\begin{multline}
  r_5 : \text{rigor}= (\text{medium} \rightarrow \text{high}) \wedge \text{grammar}= (\text{1} \rightarrow \text{0}) \Rightarrow \text{evaluation}= (\text{bad} \rightarrow \text{good} )
  \\ 
  \textit{with\:uplift\:10.82\%}. 
\label{eq:non_dominant_rule_example}  
\end{multline}
\vspace{-6mm}

For mining action rules, the Python package `action-rules`\footnote{https://github.com/lukassykora/action-rules} was used, which is based on the Action-Apriori algorithm \citep{sykora2023apriori}. This algorithm accepts action rule parameters, such as stable and flexible attributes, as well as thresholds for minimum support and confidence for both the undesired and desired parts, directly into the Apriori algorithm. This integration allows for more efficient pruning of candidate itemsets, improving the overall efficiency of the rule mining process. More detailed examples of action rules mined from the CORD-19 and M17+ datasets can be found in Section \ref{sec:action:results}.

\section{Results}\label{sec:results}

\subsection{Computational resources and complexity}
\label{sec:complexity}
This subsection outlines the computational resources and complexity of our feature extraction process. First, we describe local experiments using LLMs on various GPUs—comparing runtime, power consumption, and costs for manually generated features on the CORD-19 and M17+ datasets. Next, we detail our cloud-based approach using the OpenAI Batch API for automatic feature extraction, which leverages batch processing within a 24-hour window for cost efficiency.
\subsubsection{LLMs on local GPUs}
\label{sec:llmonlocalgpus}
The non-finetuned Llama2 13B GPTQ Chat model was run locally with the use of GPU computation as the crucial part of the experiment. 

For benchmark purposes, we generated the feature 'rigor' from the CORD-19 dataset on 2 different models utilizing 3 different GPUs. The prompt for 'rigor' is 126 tokens, while the average abstract contains 229 tokens, averaging 355 input tokens per query. We calculated the costs based on the respective GPU's worst-case power consumption, the runtime, and the average electricity price of 0.11 USD/kWh. 
The best performer, time- and money-wise, was Llama2, which was run utilizing the consumer-grade gaming GPU NVIDIA RTX 3090. DeepSeek\footnote{\url{https://huggingface.co/deepseek-ai}} took significantly longer, although it was run on the server GPU NVIDIA L40s, and at higher wattage than Llama2, which was run on entry-level server GPU NVIDIA RTX A4000.

\begin{table}[!ht]
\centering
\caption{Costs of feature extraction using different models and GPUs for determining the rigor feature value based on LLM prompt from Listing \ref{ex:llm-prompt} for 3,000 instances in CORD-19 dataset }
\renewcommand{\arraystretch}{1.5}
\begin{tabular}{l p{1cm} p{2cm} p{2.5cm} p{2.5cm}} 
\toprule
\textbf{Model@GPU} &\textbf{Max Watts} & \textbf{Time/1 call} & \textbf{Time/3000 calls} & \textbf{Price/3000 calls} \\
\midrule
Llama2@RTX 3090 & 350W & 5.91s & 4h 57min& \$0.15 \\
Llama2@A4000 & 140W & 11.35s & 9h 27min & \$0.20 \\
DeepSeek@L40s & 300W &63.72s & 53h 6min & \$1.81 \\
\bottomrule
\end{tabular}%
\label{tab:model_comparison}
\end{table}

In total, we generated 62 features for each of the 5,000 abstracts from CORD-19 and M17+ combined, resulting in 310,000 prompts required to generate the LLM features for both datasets.
\subsubsection{LLMs on cloud GPUs}
To facilitate direct comparison and assess the efficiency of cloud-based solutions, the OpenAI Batch API was employed for feature extraction across all datasets (CORD-19, M17+, BANKING77 Hate Speech and Food Hazard). The Batch API allows for the submission of extensive request batches with a guaranteed completion within a user-defined time window; in this case, a 24-hour window was selected. Opting for this 24-hour completion window resulted in a 50\% cost reduction compared to synchronous API calls. This cost efficiency is achieved by allowing OpenAI to schedule these tasks during periods of lower demand, optimizing resource utilization. Batches are submitted sequentially.

Table~\ref{tab:batch_api_costs} summarizes the costs and batch details associated with the automatic feature extraction process described in Section~\ref{sec:automatic}.

\begin{table}[!ht]
\centering
\caption{Costs of feature extraction using OpenAI Batch API (model \texttt{gpt-4o-mini-2024-07-18}).}
\renewcommand{\arraystretch}{1.5}
\begin{tabular}{lcccc}
\toprule
\textbf{Dataset} & \textbf{API Input} & \textbf{API Output} & \textbf{Total Cost} & \textbf{Batches} \\
\midrule
M17+ & \$0.36 & \$0.36 & \$0.72 & 3 \\
CORD-19 & \$0.47 & \$0.48 & \$0.96 & 4 \\
BANKING77 & \$1.31 & \$1.95 & \$3.26 & 12 \\
Hate Speech & \$1.17 & \$1.53 & \$2.71 & 11 \\
Food Hazard & \$0.66 & \$0.76 & \$1.42 & 12\\
\bottomrule
\end{tabular}%
\label{tab:batch_api_costs}
\end{table}

\subsection{LLM-augmented datasets}
\label{ss:llm-augmented-datasets}
The features generated by the language model were added to all CORD-19, M17+, BANKING77, Hate Speech, and Food Hazard datasets. Ordinal features such as \textit{rigor} and \textit{novelty} for CORD-19 and M17+ were ordinally encoded. 

For the user-selected features extracted with local LLM inference, only 2 articles in the CORD-19 dataset were assigned feature values outside of the allowed feature space. These instances were removed, making the final length of CORD-19 2,998 articles (instances).  All LLM-generated feature values for the M17+ articles were valid, resulting in a dataset of 2,000 instances. 
The first few rows from both generated datasets are included in Table~\ref{tab:c19_table_sample} and Table~\ref{tab:m17_table_sample}.

All rows generated in the automatic workflow by gpt-4o-mini with Batch API were valid. A sample of the BANKING77 LLM-generated features can be seen in Table~\ref{tab:bank77_table_sample}.
\begin{table}[!ht]
\centering
\resizebox{\textwidth}{!}{\begin{tabular}{lllcllccc}
  \hline
  abstract & rigor & novelty & grammar & accessibility & math & compsci & ... & target\\ 
  \hline
Family... & medium & medium & 0 & high & 0 & 0 & ...  & 0\\  
Patients... & medium & low & 0 & medium & 0 & 0 & ... & 0\\  
Coronavirus... & high & high & 0 & medium & 0 & 0 & ... & 1\\  
We present... & high & high & 1 & high & 0 & 1 & ...  & 1\\ 
Gender... & low & high & 0 & high & 0 & 0 & ...  & 0\\  
   \hline
\end{tabular}}
\caption{Sample of enriched CORD-19 dataset with added LLM-generated feature values (for user-specified features). Target is a binary variable - with 1 representing works cited more than the median value adjusted in time, 0 represents articles with lower or equal to the median citation count (details in \cite{Beranova2022}). }
\label{tab:c19_table_sample}
\end{table}

\begin{table}[!ht]
\centering
\resizebox{\textwidth}{!}{\begin{tabular}{lllcllccc}
  \hline
  abstract & rigor & novelty & grammar & accessibility & math & compsci & ...  & target\\ 
  \hline
This... & high & medium & 0 & medium & 0 & 0 & ... & 1\\  
Introduction... & medium & medium & 1 & medium & 0 & 1 & ...& 2\\  
The present... & medium & low & 1 & medium & 0 & 0 & ... & 4\\  
The main... & low & low & 1 & low & 0 & 1 & ... & 5\\ 
Previous... & medium & low & 0 & medium & 0 & 0 & ...   & 3\\  
   \hline
\end{tabular}}
\caption{Sample of enriched M17+ dataset with added LLM-generated feature values (for user-specified features). Target is an ordinal variable of human evaluation of the quality of articles, 1 indicates works of world-class quality, 5 means mediocre articles of low importance and impact.}
\label{tab:m17_table_sample}
\end{table}

\begin{table}[!ht]
\centering
\resizebox{\textwidth}{!}{\begin{tabular}{lcccllccc}
  \hline
  text & contains\_question & contains\_card\_mention & contains\_currency & intent\_category  & ...  & target\\ 
  \hline
I am still... & no & yes & no & activate\_my\_card  &  ... & 11\\  
What can I... & yes & yes & no & activate\_my\_card &  ...& 11\\  
Can I link... & yes & yes & no & activate\_my\_card &  ... & 13\\  
There is a... & yes & no & yes & unknown\_charge  &  ... & 34\\ 
Which fiat... & yes & yes & yes & wrong\_exchange\_rate &  ...   & 36\\  
   \hline
\end{tabular}}
\caption{Sample of enriched BANKING77 dataset with added LLM-generated feature values (for automatically generated features). The target variable represents the 77 distinct customer intents.}
\label{tab:bank77_table_sample}
\end{table}

\subsection{Validating LLM features}
This subsection documents the validation of LLM-generated features, which was conducted with the help of three techniques. First, the features from each dataset were, along with their dataset's respective target variable, tested for a statistically significant relationship. This was done on 2,500 bootstrapped samples from the original datasets. Furthermore, Cramér's V was computed to document the strength of the association of these features and the target of the respective variables. Two selected LLM-generated features were compared against features that were extracted using non-LLM techniques. Finally, we report on a user study aimed at human evaluation of automatically discovered features, which complements the other feature evaluations with a new perspective. While the users may not be able to precisely identify feature quality, they are able to assess whether features subjectively appear as relevant or not.

\subsubsection{Analysis of relationship between target and LLM-features}
\label{ss:responseto8}
To formally test whether generated features have a relationship with the respective target variables, we employed the chi-squared test of independence at significance level $\alpha = 0.05$. We further performed this test on 2,500 bootstrapped samples for each feature to verify the results.

\subsubsection{Effect sizes}
To validate the results obtained from the chi-squared test and ensure that the statistically significant associations are not due to large sample sizes, we tested the effect size using Cramér's V \citep{cramér1946mathematical}, see Equation \ref{eq:cramersV}.

\begin{equation}
V = \sqrt{\frac{\chi^2}{n \cdot (k-1)}},
\label{eq:cramersV}
\end{equation}
where $\chi^2$ is the chi-squared statistic, $n$ is the sample size and $k$ is the number of categories in the smaller dimension of the contingency table.

\begin{table}[!ht]
\centering
\small
\caption{Sample of statistically significant features according to the relation between \textit{target} and LLM-generated features (based on user-specified feature list), dataset \textbf{M17+}. The star notation indicates the level of statistical significance: $^{***}p < 0.001$, $^{**}p < 0.01$, and $^{*}p < 0.05$. }
\begin{tabular}{l p{0.9cm} l | l p{0.9cm} l}
\hline
Feature & Imp. & Cramér's V & Feature & Imp. & Cramér's V \\
\hline
BioSci\footnotemark[2]        & *** & 0.349 & Replicability\footnotemark[1]        & *** & 0.322 \\
Basic Medicine\footnotemark[2] & *** & 0.314 & PhysSci\footnotemark[2]             & *** & 0.305 \\
Rigor\footnotemark[1]          & *** & 0.296 & OtherNatSci\footnotemark[2]         & *** & 0.293 \\
CompSci\footnotemark[2]        & *** & 0.280 & OtherMedSci\footnotemark[2]         & *** & 0.265 \\
ChemSci\footnotemark[2]        & *** & 0.261 & MedBiotech\footnotemark[2]          & *** & 0.257 \\
AnimalSci\footnotemark[2]      & *** & 0.250 & VeterinarySci\footnotemark[2]       & *** & 0.248 \\
Psychology\footnotemark[2]     & *** & 0.246 & Novelty\footnotemark[1]             & *** & 0.245 \\
EarthSci\footnotemark[2]       & *** & 0.243 & OtherAgrolBiotech\footnotemark[2]    & *** & 0.234 \\
Clinical Medicine\footnotemark[2] & *** & 0.228 & Grammar\footnotemark[1]           & *** & 0.220 \\
EnviroBiotech\footnotemark[2]  & *** & 0.218 & OtherSocSci\footnotemark[2]         & *** & 0.213 \\
HealthSci\footnotemark[2]      & *** & 0.208 & Agriculture\footnotemark[2]         & *** & 0.204 \\
Math\footnotemark[2]           & *** & 0.196 & IndustBiotech\footnotemark[2]        & *** & 0.191 \\
Mixed methods\footnotemark[3]  & *** & 0.186 & LangLiter\footnotemark[2]           & *** & 0.179 \\
\hline
\end{tabular}
\footnotetext[1]{Categorical features based on abstract quality and contents.}
\footnotetext[2]{Binary features belonging to \textit{disciplines} values.}
\footnotetext[3]{Binary features belonging to \textit{research} values.}
\label{tab:m17_test}
\end{table}

We present detailed significance testing results in Table \ref{tab:m17_test} for M17+, which we chose to represent scientometric datasets. Higher values of Cramér's V are observed for features in the M17+ dataset and target variable. 
The M17+ dataset has a larger number of features that indicate a statistically significant relationship between the LLM-generated features and the awarded grade. For example, \textit{grammar}, \textit{replicability} all have a statistically significant relationship with the target value.
Results for Food Hazard, which we chose as the representative of the other datasets, are included in Table~\ref{tab:foodhazard_test}. The most important features are \textit{company\_name} and \textit{hazard\_type}. The latter is the LLM's attempt at assessing the \textit{hazard\_category} target variable, which works well. Other features important for hazard category classification are \textit{recall\_reason}, \textit{contaminant\_type} or \textit{recall\_severity}.

\begin{table}[!ht]
\centering
\small
\caption{Sample of statistically significant features according to the relation between \textit{hazard-category} and LLM-generated features (based on LLM-generated feature list), dataset \textbf{Food Hazard}. The star notation indicates the level of statistical significance: $^{***}p < 0.001$, $^{**}p < 0.01$, and $^{*}p < 0.05$.}
\begin{tabular}{l p{0.9cm} l | l p{0.9cm} l}
  \hline
  Feature & Imp. & Cramér's V & Feature & Imp. & Cramér's V \\
  \hline
  company\_name    & *** & 0.865 & hazard\_type             & *** & 0.817 \\
  recall\_reason   & *** & 0.582 & contaminant\_type        & *** & 0.448 \\
  recall\_severity & *** & 0.300 & allergen\_type           & *** & 0.274 \\
  recall\_source   & *** & 0.227 & product\_origin          & *** & 0.204 \\
  product\_type    & *** & 0.164 & distribution\_area       & *** & 0.158 \\
  recall\_frequency& *** & 0.154 & product\_packaging       & *** & 0.119 \\
  consumer\_advice & *** & 0.116 & affected\_population     & *** & 0.101 \\
  recall\_action   & *** & 0.097 & recall\_notification\_method & *** & 0.088 \\
  recall\_duration & *** & 0.087 &                          &     &       \\
  \hline
\end{tabular}
\label{tab:foodhazard_test}
\end{table}


\subsubsection{Validating LLM-features against domain-specific feature generation}
\label{sec:llm-feature-validation }
To help evaluate the quality of LLM-generated features, we applied non-LLM-based extraction methods to obtain  'grammar' and 'accessibility' features for the M17+ dataset. These are then compared with the same features whose values were derived through an LLM.

For 'grammar', the LanguageTool \citep{languagetool} grammar checker was used through the Python wrapper tool python-language-tool \citep{morris2025languagetool}. \par
The LLM-generated feature 'accessibility' as defined in Table \ref{tab:abstract_criteria} was compared with the SMOG Index \citep{d9397c09-9d7e-3784-b191-6efaa0fd35d0}, which is defined in Equation \ref{eq:smogindex} and is commonly used to evaluate the readability of text and provides the approximate education level required to understand the provided text. \par

\begin{equation}
\text{SMOG Index} = 1.043 \times \sqrt{ \frac{\text{Polysyllabic Words} \times 30}{\text{Number of Sentences}} } + 3.1291
\label{eq:smogindex}
\end{equation}
\vspace{4mm}

For the 'accessibility' feature, we obtained $\rho_{accessibility, SMOG Index} = 0.115$, indicating a weak positive correlation between the features. This goes against our hypothesis that article abstracts marked as more accessible will be awarded a lower SMOG Index. This is likely due to the fact that the SMOG Index uses the number of polysyllabic words per sentence as the measure of readability, while the prompt for the extraction of the 'accessibility' feature instructed the language model to focus on the clarity of language and how easily the abstracts communicate the main points of the research. Furthermore, the following results were calculated: $\rho_{accessibility, evaluation} = -0.06$ and $\rho_{SMOG Index, evaluation} = -0.199$, meaning that the SMOG Index correlates with our target variable on this dataset more than the LLM-generated feature 'accessibility'. 

In the case of the 'grammar' feature, the correlation coefficient $\rho = 0.141$ indicates a weak correlation between the LLM-generated and LanguageTool assessments. However, when we evaluated the correlations between the target variable and the respective features, we found that the LLM-generated grammar assessment correlates with the target variable with $\rho_{grammar, evaluation} = 0.218$ more than the LanguageTool-based feature, which showed no significant correlation with $\rho_{grammar\_ltp, evaluation} = -0.004$. 

These findings suggest that while LLM-generated features may not perfectly mirror traditional metrics like the SMOG Index or LanguageTool outputs (showing weak direct correlations), they can capture aspects relevant to the target outcome. This is evidenced by the LLM 'grammar' feature correlating more strongly with the expert 'evaluation' score than the standard tool's output, despite the mixed results for the 'accessibility' feature.

\subsubsection{Predictive performance -  LLM features based on user set feature list (scientometric datasets)}
\label{sec:performance_user}
In the classification task on respective datasets - binary classification on CORD-19 and ordinal classification on M17+, we tested various models on multiple datasets and chose the best models based on test accuracy. For the purposes of training the machine-learning models, we did not downsample the feature space to only the statistically significant features and used all LLM-generated features.

As the M17+ dataset contains an ordinal target variable, it is also useful to track the mean absolute error (MAE), which provides information about the average error of the predicted evaluation. We also report the test F1 scores, which do not differ greatly from the accuracy score, as both datasets were balanced in previous steps. 
Table~\ref{tab:combined_metrics} reports three crucial model performance metrics - test accuracy, F1 score, and MAE.

For CORD-19, the \textit{LLM-feature only} set showed significant improvement over the naive baseline model, indicating a 9 percentage point (p.p.) improvement. Furthermore, it scored only 3 p.p. lower than the state-of-the-art SciBERT embeddings for scientific text. The combination of LLM-generated features with TF-IDF terms further improved the test accuracy and F1 score, surpassing both SciBERT embeddings and TF-IDF. 

The M17+ features a 5-class ordinal variable, making the task of text classification significantly harder, which is why the overall scores are generally worse for this dataset. Nevertheless, the prediction of human evaluation on the M17+ showed a more pronounced change when compared to the naive baseline model, increasing the test accuracy and F1 score almost twofold. The best-performing model trained on the \textit{LLM-features only} set was Gradient Boost Classifier, which surpassed the TF-IDF based model in terms of test accuracy and F1 score and reached results comparable to the state-of-the-art black box AutoGluon model. 
None of the acquired models reached the results achieved by the model trained on SciBERT embeddings, which scored best in all 3 observed metrics but has the main downside of providing no interpretability to its encoded features. 

\begin{table}[!ht]
    \caption{Comparison of Accuracy, F1 Score, and MAE for models trained on each dataset variation, CORD-19 (C19) and M17+. User-selected features only.}
    \centering
    \small
    \renewcommand{\arraystretch}{1.2} 
    \begin{tabular}{p{4cm}p{1cm}p{1cm}p{1cm}p{1cm}p{1cm}p{1cm}}
        \toprule
        & \multicolumn{2}{c}{\textbf{Accuracy}} & \multicolumn{2}{c}{\textbf{F1 Score}} & \multicolumn{2}{c}{\textbf{MAE}} \\
        \cmidrule(lr){2-7}
        \textbf{Model} & \textbf{C19} & \textbf{M17+} & \textbf{C19} & \textbf{M17+} & \textbf{C19} & \textbf{M17+} \\
        \midrule
        text + LLM features (AutoGluon) & 0.665 & 0.395 & 0.664 & 0.389 & 0.335 & 0.855 \\
        BoW + LLM-features  & 0.653 & 0.393 & 0.653 & 0.377 & 0.347 & 1.068 \\
        text only (AutoGluon) & 0.630 & 0.377 & 0.628 & 0.324 & 0.370 & 1.001 \\
        SciBERT embeddings     & 0.625 & 0.408 & 0.625 & 0.392 & 0.335 & 0.848 \\
        TF-IDF  & 0.625 & 0.343 & 0.622 & 0.332 & 0.375 & 1.073 \\
        LLM-features & 0.597 & 0.355 & 0.597 & 0.326 & 0.403 & 1.113 \\
        Naive classifier & 0.502 & 0.180 & 0.502 & 0.180 & 0.499 & 1.620 \\
        \bottomrule
    \end{tabular}

    \label{tab:combined_metrics}
\end{table}

\subsubsection{Predictive performance - features selected with LLMs}
\label{ss:predictive-performance-features}
LLM-generated features that were first selected by an LLM were employed in a classification task and evaluated in the same way as user-selected features in the previous Section \ref{sec:performance_user}. 

The choice of the evaluation measures in Tables~\ref{tab:metrics_m17_c19_gpt}-\ref{tab:combined_metrics_llm_select} reporting the results depends on the type of the target variable in the dataset.
Datasets  CORD-19 and M17+ have ordinal/binary targets, while  BANKING77, Hate Speech, and Food Hazard datasets have multinomial targets.

The BANKING77 consists of 77 equally represented target classes. The model employing only the LLM-generated features showed an increase in all metrics over the naive uniform classifier, which performed poorly as expected. The same model did not reach the test classification metrics of the TF-IDF-based model. However, combining the LLM-generated features with TF-IDF terms resulted in, albeit a minor, increase in observed metrics over the TF-IDF-only approach.

The model trained on the LLM-feature-only features subset for Hate Speech dataset showed improvement in all metrics besides accuracy when compared to the naïve classifier, which is expected due to class imbalance. Furthermore, while behind in accuracy and precision when compared to the TF-IDF term-based classifier, it showed a 13 p.p. improvement in recall. Upon further inspection, we learned that the class recall of the '1' class (hate speech present) increased from 0.14 in the TF-IDF to 0.54 for the LLM-feature only set and to 0.35 for the TF-IDF + LLM-feature set. This result is particularly important as more hate speech comments were successfully detected with LLM-generated features as opposed to the TF-IDF model.

For the Food Hazard dataset, we focused on predicting the 'hazard-category' target variable, which is one of the tasks of sub-task 1 of the Semeval challenge. Results showed that the LLM-feature only set increased recall over both TF-IDF only and TF-IDF + LLM feature sets while having lower precision. This demonstrates the LLM-feature-based model's ability to recognize more niche food hazard instances than the traditional TF-IDF-based approach. 

Furthermore, the features selected by the LLMs only on the CORD-19, and the M17+ datasets were used in a similar classification experiment to the user-selected features. When directly comparing the selected metric values from Table~\ref{tab:combined_metrics} and Table~\ref{tab:metrics_m17_c19_gpt}, we can see that both user features and LLM-selected features generated for CORD-19 performed similarly, while user-selected features outperformed the LLM-selected features slightly on the M17+ dataset.

\begin{table}[!ht]
    \caption{Comparison of Accuracy, F1 Score, and MAE for models trained on each dataset variation, CORD-19 (C19) and M17+. Features selected with LLMs only.}
    \centering
    \small
    \renewcommand{\arraystretch}{1.2} 
    \begin{tabular}{p{4cm}p{1cm}p{1cm}p{1cm}p{1cm}p{1cm}p{1cm}}
        \toprule
        & \multicolumn{2}{c}{\textbf{Accuracy}} & \multicolumn{2}{c}{\textbf{F1 Score}} & \multicolumn{2}{c}{\textbf{MAE}} \\
        \cmidrule(lr){2-7}
        \textbf{Model} & \textbf{C19} & \textbf{M17+} & \textbf{C19} & \textbf{M17+} & \textbf{C19} & \textbf{M17+} \\
        \midrule
        TF-IDF  & 0.63 & 0.34 & 0.62 & 0.33 & 0.38 & 1.07 \\
        LLM-features & 0.59 & 0.33 & 0.59 & 0.33 & 0.41 & 1.17 \\
        Naïve classifier & 0.50 & 0.18 & 0.50 & 0.18 & 0.50 & 1.62 \\
        \bottomrule
    \end{tabular}

    \label{tab:metrics_m17_c19_gpt}
\end{table}

\begin{table}[!ht]
    \caption{Comparison of Accuracy, Recall, and Precision for models trained on each dataset variation: BANKING77 (B77), Hate Speech (Hate), and Food Hazard (Haz). Naïve classifiers were chosen according to the target class characteristics: uniform classifier for BANKING77 and most frequent class for Hate Speech and Food Hazard.}
    \centering
    \small
    \renewcommand{\arraystretch}{1.2} 
    \begin{tabular}{p{3.6cm}*{9}{p{0.5cm}}}
        \toprule
         & \multicolumn{3}{c}{\textbf{Accuracy}} & \multicolumn{3}{c}{\textbf{Precision}} & \multicolumn{3}{c}{\textbf{Recall}} \\
        \cmidrule(lr){2-4} \cmidrule(lr){5-7} \cmidrule(lr){8-10}
        \textbf{Model} & \textbf{B77} & \textbf{Hate} & \textbf{Haz} & \textbf{B77} & \textbf{Hate} & \textbf{Haz} & \textbf{B77} & \textbf{Hate} & \textbf{Haz} \\
        \midrule
        TF-IDF                & 0.77 & 0.87 & 0.91 & 0.78 & 0.60 & 0.80 & 0.77 & 0.39 & 0.55 \\
        LLM-features      & 0.59 & 0.65 & 0.93 & 0.64 & 0.35 & 0.77 & 0.64 & 0.52 & 0.64 \\
        Naïve classifier      & 0.01 & 0.87 & 0.37 & 0.01 & 0.25 & 0.10 & 0.01 & 0.22 & 0.05 \\
        \bottomrule
    \end{tabular}
    \label{tab:combined_metrics_llm_select}
\end{table}

\subsubsection{Explanation of generated features}
The explanation of features with the use of Shapley values was done on two pairs of classifiers. First, we compared a classifier trained on the results of a fully automated LLM-based feature generation with a classifier that was trained on user-selected features combined with  LLM-based feature generation. This evaluation was done on the M17+ dataset. To complement this, we present the Shapley values for two additional datasets using the automated LLM feature extraction setup.  

\begin{figure}[!ht]
    \caption{SHAP plot for a classification model trained on \textit{LLM-feature only} feature set for the M17+ dataset:  automated LLM-based feature discovery combined with LLM-based feature generation (left) vs user-selected features combined with LLM-based feature generation  (right). The X-axis represents the SHAP value, indicating the feature's effect on the prediction—positive values increase the prediction, while negative values decrease it. Features are ranked by importance on the Y-axis. The color of the points corresponds to the feature values. For binary features like grammar or basic\_medicine, blue indicates a value of 0, while red represents a value of 1. For categorical features such as rigor or novelty, blue denotes a low value, purple indicates a medium value, and red signifies a high value.}
    \centering
    \includegraphics[scale=0.65]{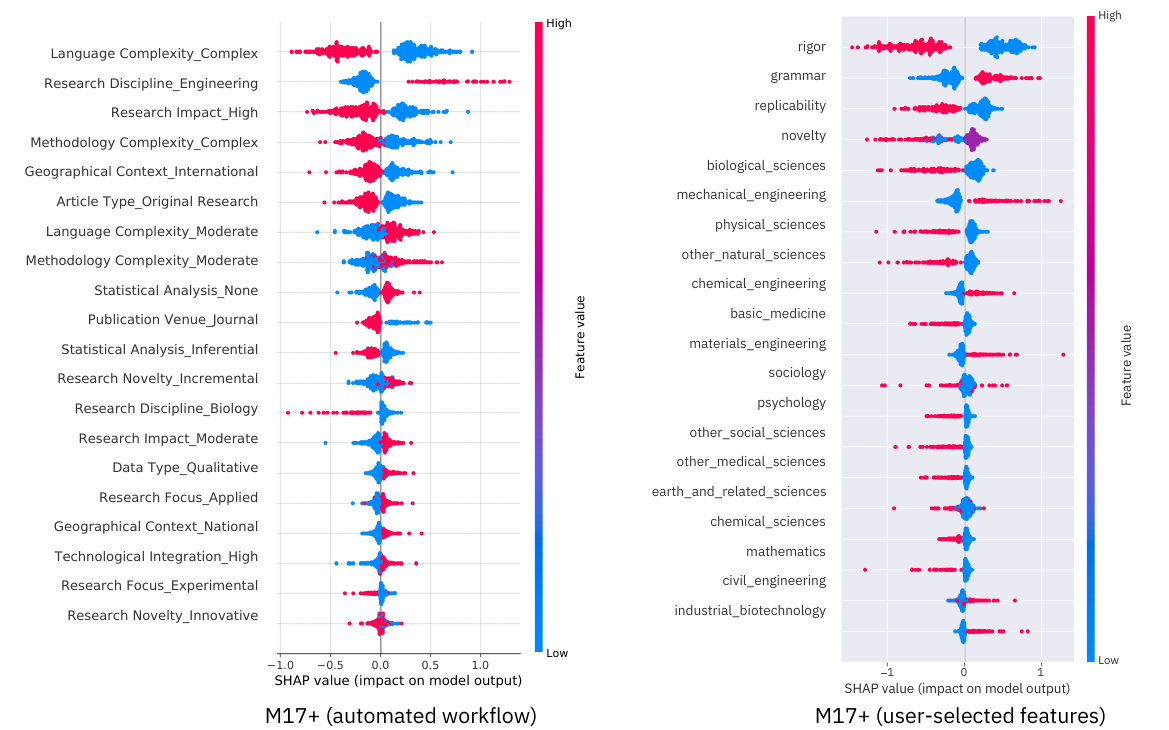}
    \label{fig:shap-manual-vs-auto-m17} 
\end{figure}

Results shown in Figure~\ref{fig:shap-manual-vs-auto-m17} indicate that for the M17+ dataset, the most important features from the user-selected ones were \textit{rigor}, \textit{grammar}, \textit{replicability}, and \textit{novelty}. For a classifier trained on fully automatic feature extraction, the most important features are language complexity, research discipline engineering, research impact high, and complex methodology. The comparison of both results shows a match as the ``manual'' grammar feature is closely related to ``automatic'' language complexity, rigor is related to the complexity of methodology, and novelty is related to high research impact.  

Interestingly, both analyses using SHAP values support our initial hypothesis by showing that higher novelty is linked to better expert evaluations. This association is further validated by the statistical tests and effect sizes discussed in previous sections.

\begin{figure}[!ht]
    \caption{SHAP plot for classification models trained on \textit{LLM-feature only} feature set for the Hate Speech dataset, target class 1 (hate speech present) (left) and the Food Hazard dataset target class 1 (biological food hazard) (right). For guidance on the general interpretation of the SHAP plot, refer to the caption of the previous figure.}
    \centering
    \includegraphics[scale=0.65]{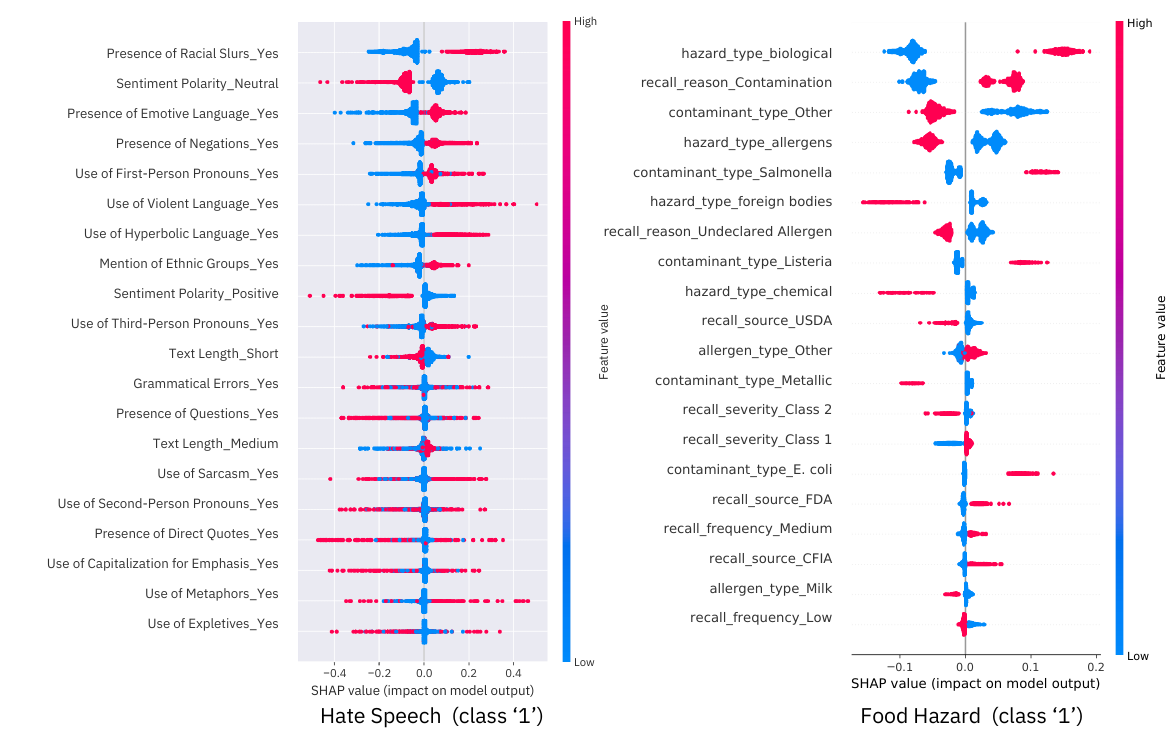}
    \label{fig:shap-hazard-vs-hate} 
\end{figure}

In Figure \ref{fig:shap-hazard-vs-hate} (left), the analysis shows that for the Hate Speech dataset and specifically for predicting class '1' (the presence of hate speech), the feature \textit{Presence of Racial Slurs\_Yes} has the highest impact on the model's output. Additionally, the SHAP values indicate that non-neutral sentiment and the use of emotive language are strongly associated with hate speech, as could be expected. 

Figure \ref{fig:shap-hazard-vs-hate} (right) shows that the most important LLM-generated feature is the LLM's attempt to assess the target label for the Food Hazard dataset directly, as we can see from the feature \textit{hazard\_type\_biological}. It can also be observed that the presence of Salmonella and the stated recall reason being contamination are intuitively understandable as important features, also for non-experts.

\subsubsection{User-perceived relevance of LLM-discovered features}
To complement the quantitative evaluation of automatically discovered features by evaluating their predictive performance and match with domain-specific feature generation methods, we also conducted a small-scale study of human perception of the generated features.

\begin{figure}[!ht]
  \centering
  \begin{subfigure}[t]{0.48\textwidth}
    \centering
    \includegraphics[width=\linewidth]{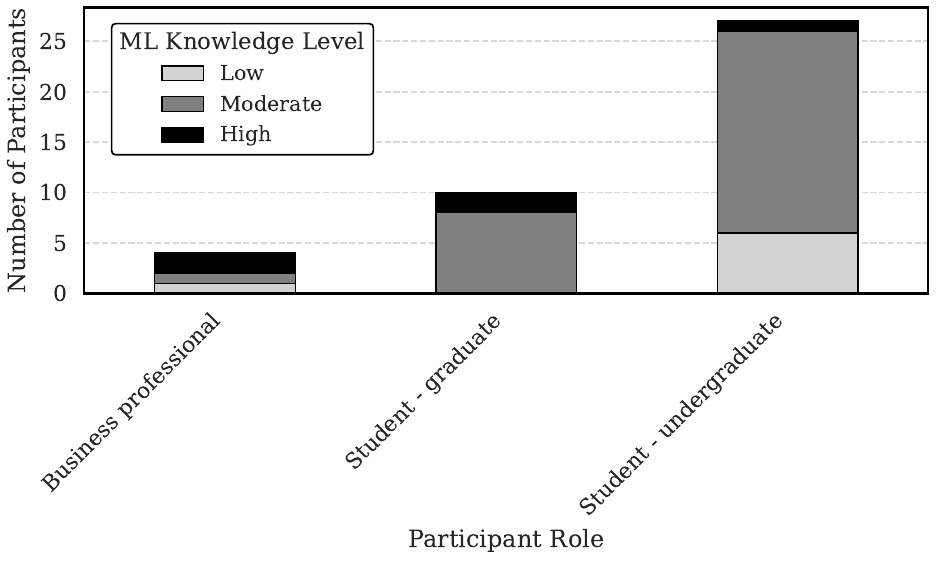}
    \caption[Participant demographics]{Survey participants ($N = 41$) by role and machine learning knowledge.}
    \label{fig:sub1}
  \end{subfigure}%
  \hspace{0.02\textwidth}%
  \begin{subfigure}[t]{0.48\textwidth}
    \centering
    \includegraphics[width=\linewidth]{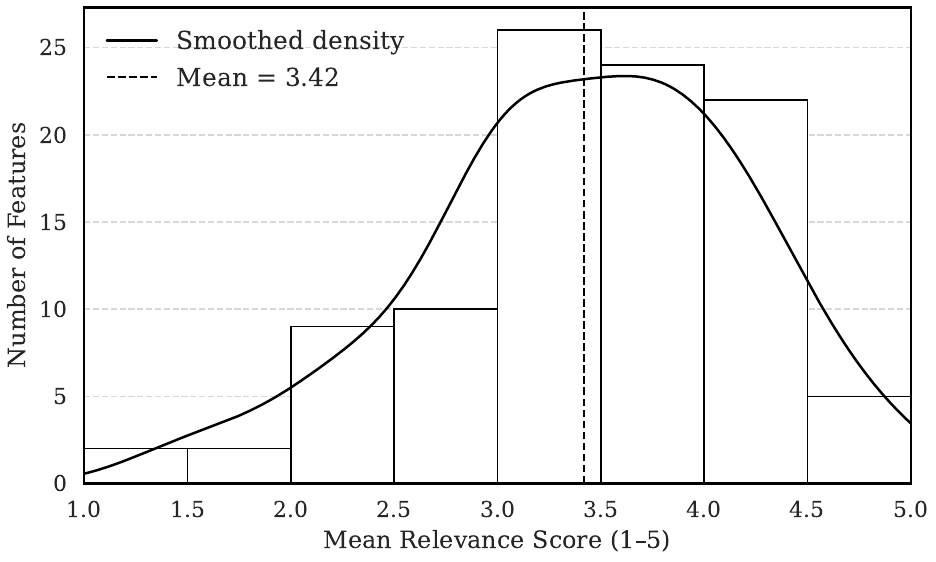}
    \caption[Feature relevance histogram]{Histogram of mean relevance scores (1 = “Not relevant” to 5 = “Relevant”) for N=100 evaluated features pooled from 5 datasets.}
    \label{fig:sub2}
  \end{subfigure}
  \caption{%
Survey of user-perceived relevance of automatically discovered features. Each survey participant rated the relevance of 20 features that were automatically discovered for one of five datasets (CORD-19, M17+, BANKING77, Hate Speech and Food Hazard).}
  \label{fig:combined}
\end{figure}

To evaluate the appropriateness of automatically discovered features, we surveyed \(41\) respondents. Participants self-identified as Business Professionals, Academic/Researchers, Undergraduate Students, or Graduate Students, and reported their machine learning expertise as None, Low (minimal understanding, no practical use), Moderate (understands core concepts with limited practice), or High (strong theoretical/practical experience), as shown in Figure~\ref{fig:sub1}.

Each participant was randomly presented a brief description of one of five distinct datasets and its corresponding prediction task, for which they were asked to rate the relevance of 20 automatically discovered features.
Similarly to \cite{heo2022comparison}, we asked the participants to rate feature relevance on a 5‑point relevance scale (1 = “Not relevant” to 5 = “Relevant”).

The distribution of results, shown in Figure~\ref{fig:sub2}, is positively skewed toward higher values. Of the 100 features that were evaluated in total across the five datasets, only four features were considered as clearly not relevant (mean score between 1.0 and 2.0). 
In contrast, 27 features were evaluated as clearly relevant (mean score 4.0 and 5.0). Similarly, 50 features had a mean relevance score between 3.0 and 4.0, while 19 had a mean score between 2.0 and 3.0.  The low number of features considered as not relevant suggests that LLM rarely outputs completely irrelevant or ``hallucinated'' features. With a substantial proportion of discovered features being judged as relevant by the evaluators, the results support the efficacy of our automatic generation pipeline.

\subsection{Case Study: Action Rule Mining for Article Improvement}\label{sec:action:results}
We focus on applying action rules on scientometric datasets as these present the most promising opportunities for interventions proposed by action rules. For the analysis in this section, we used the CORD-19 and M17+ dataset versions with user-defined features.

For action rule mining, attributes are divided into two categories: stable attributes, which remain unchanged and serve as fixed conditions in the rules, and flexible attributes, where changes are pursued to achieve the desired outcome. The attribute \textit{area} is classified as stable, while \textit{novelty}, \textit{rigor}, \textit{grammar}, \textit{replicability}, \textit{accessibility}, and \textit{research} are categorized as flexible attributes. This classification is used as input in the \texttt{action-rules} Python package and is consistent across both datasets. Four experiments are conducted, and the settings for the \texttt{action-rules} package are detailed in Table \ref{tab:action_rules_experiments}. In addition to the settings, this table also presents the number of discovered action rules and the count of dominant action rules derived from them.

\begin{table}[!ht]
    \centering
    \small
    \renewcommand{\arraystretch}{1.2} 
    \begin{tabular}{p{4cm}p{1.6cm}p{1.6cm}p{1.6cm}p{1.6cm}}
        \toprule
        \textbf{Settings / Experiments} & \textbf{\ref{sec:experiment1}} & \textbf{\ref{sec:experiment2}} & \textbf{\ref{sec:experiment3}} & \textbf{\ref{sec:experiment4}} \\
        \midrule
        Dataset & M17+ & M17+ & M17+ & CORD-19 \\
        Min. Stable Attributes & 0 & 0 & 1 & 0 \\
        Min. Flexible Attributes & 1 & 1 & 1 & 1 \\
        Min. Undesired Support & 60 & 30 & 35 & 20 \\
        Min. Desired Support & 60 & 40 & 35 & 20 \\
        Min. Undesired Conf. & 70\% & 70\% & 70\% & 70\% \\
        Min. Desired Conf. & 70\% & 70\% & 70\% & 70\% \\
        Target & Evaluation & Evaluation & Evaluation & Cited \\
        \hspace{0.4cm}Undesired State & Bad & Avg & Bad & 0 \\
        \hspace{0.4cm}Desired State & Good & Best & Good & 1 \\
        \arrayrulecolor{lightgray}\midrule
        Discovered Action Rules & 42 & 22 & 21 & 16 \\
        Dominant Action Rules & 6 & 5 & - & 6 \\
        \arrayrulecolor{black}\bottomrule
    \end{tabular}
    \caption{Overview of action rule settings and results across different experiments (experiment ids correspond to paper section numbers).}
    \label{tab:action_rules_experiments}
\end{table}

\subsubsection{M17+ Dataset - Bad to Good Evaluation Transition} \label{sec:experiment1}

\begin{table}[!ht]
    \centering
    \small
    \renewcommand{\arraystretch}{1.2} 
    \begin{tabular}{p{0.3cm}p{1cm}p{1.7cm}p{1.7cm}p{0.8cm}p{0.8cm}p{0.4cm}p{1.7cm}p{0.8cm}}
        \toprule
        \rotatebox{90}
        {\textbf{Rule}} & \rotatebox{90}{\textbf{Area}} & \rotatebox{90}{\textbf{Novelty}} & \rotatebox{90}{\textbf{Rigor}} & \rotatebox{90}{\textbf{Grammar}} & \rotatebox{90}{\textbf{Replicability}} & \rotatebox{90}{\textbf{Accessibility}} & \rotatebox{90}{\textbf{Evaluation}} & \rotatebox{90}{\textbf{Uplift}} \\
        \midrule
        $r_6$ & & & med $\rightarrow$ high & & 0 $\rightarrow$ 1 & & bad $\rightarrow$ good & 16.96\% \\
        $r_7$ & & & med $\rightarrow$ high & & & & bad $\rightarrow$ good & 16.95\% \\
        $r_8$ & & & & 1 $\rightarrow$ 0 & 0 $\rightarrow$ 1 & & bad $\rightarrow$ good & 11.26\% \\
        $r_9$ & & low $\rightarrow$ high & & & 0 $\rightarrow$ 1 & & bad $\rightarrow$ good & 6.73\% \\
        $r_{10}$ & & low $\rightarrow$ high & & 1 $\rightarrow$ 0 & & & bad $\rightarrow$ good & 4.36\% \\
        $r_{11}$ & & low $\rightarrow$ high & & 1 & & & bad $\rightarrow$ good & 3.96\% \\
        \arrayrulecolor{lightgray}\midrule
        $r_{12}$ & & med $\rightarrow$ high & med $\rightarrow$ high & & & & avg. $\rightarrow$ best & 4.57\% \\
        $r_{13}$ & & med $\rightarrow$ high & & 1 $\rightarrow$ 0 & & & avg. $\rightarrow$ best & 3.58\% \\
        $r_{14}$ & & low $\rightarrow$ high & med $\rightarrow$ high & & & & avg. $\rightarrow$ best & 2.24\% \\
        $r_{15}$ & & low $\rightarrow$ high & & & 0 $\rightarrow$ 1 & & avg. $\rightarrow$ best & 1.95\% \\
        $r_{16}$ & & low $\rightarrow$ high & & 1 $\rightarrow$ 0 & & & avg. $\rightarrow$ best & 1.24\% \\
        \arrayrulecolor{lightgray}\midrule
        $r_{17}$ & compu. & med $\rightarrow$ high & med $\rightarrow$ high & & & & bad $\rightarrow$ good & 1.98\% \\
        $r_{18}$ & elect. & & med $\rightarrow$ high & & & med & bad $\rightarrow$ good & 1.15\% \\
        $r_{19}$ & mechan. & med $\rightarrow$ high &  & 1 $\rightarrow$ 0 & 0 $\rightarrow$ 1 & & bad $\rightarrow$ good & 1.51\% \\
        $r_{20}$ & chemic. & med $\rightarrow$ high & med $\rightarrow$ high & & & & bad $\rightarrow$ good & 1.88\% \\
        $r_{21}$ & mater. & med $\rightarrow$ high & & & 0 $\rightarrow$ 1 & & bad $\rightarrow$ good & 1.54\% \\
        \arrayrulecolor{black}\bottomrule
    \end{tabular}
    \caption{Overview of the action rules discovered from the M17+ dataset and their respective uplift percentages.}
    \label{tab:action_rules_uplift}
\end{table}

The first dataset analyzed in this study is M17+. Action rules are identified within the dataset to find those capable of changing the evaluation of articles from bad (rated 4 and 5) to good (rated 1 and 2). A total of 42 action rules are discovered. The rule with the highest impact, as shown in Equation \ref{eq:action_rule_r6}, modifies the classification of 16.96\% of articles in the dataset.

\vspace{-8mm}
\begin{multline}
  r_6 : \text{rigor}= (\text{med} \rightarrow \text{high}) \wedge \text{replicability}= (\text{0} \rightarrow \text{1}) \Rightarrow \text{evaluation}= (\text{bad} \rightarrow \text{good} )
  \\ 
  \textit{with\:uplift\:16.96\%}. 
\label{eq:action_rule_r6}  
\end{multline}
\vspace{-6mm}

Table \ref{tab:action_rules_uplift} presents the action rule $r_6$ alongside the other discovered dominant action rules. Notably, the first three rules ($r_6$, $r_7$, and $r_8$) demonstrate the potential to improve the classification of more than 10\% of articles to a better evaluation. The dominant action rules identified from this experiment are rules $r_6$ through $r_{11}$, all of which meet the settings outlined in Table \ref{tab:action_rules_experiments}.

\subsubsection{M17+ Dataset - Average to Best Evaluation Transition} \label{sec:experiment2}

Action rules are identified that have the potential to transform articles with an average evaluation (rated 3) into articles with the best evaluation (rated 1). In this analysis, 22 action rules are discovered. 

Among these 22 action rules, 5 dominant rules are identified, which are listed in Table \ref{tab:action_rules_uplift}. These dominant rules correspond to rules $r_{12}$ through $r_{16}$. In this case, the potential impact of the rules is notably reduced compared to the first experiment, with none of the rules achieving an uplift greater than 5\%. This suggests that improving an article from average to excellent is a more challenging task than improving it from poor to good.

\subsubsection{M17+ Dataset - Area-Specific Evaluation Improvement} \label{sec:experiment3}

The action rules with the highest uplift for each area are presented. In this experiment, the parameter \textit{min. stable attributes} is set to 1, meaning that the discovered rules are fixed to specific areas. The rules discovered through this experiment are $r_{17}$ to $r_{21}$. The focus of this experiment is again on identifying changes that could improve the evaluation of articles from bad (rated 4 and 5) to good (rated 1 and 2). One rule with the highest uplift is selected for each area and included in Table \ref{tab:action_rules_uplift}. In this case, dominant action rules are not considered. As anticipated, the uplift is lower, which is attributed to the fixation on specific areas, significantly reducing the number of articles to which the recommended actions apply. There are also areas for which no rules were found, and this could be addressed by adjusting the support and confidence thresholds to lower values. 

Rules $r_6$ and $r_7$ stand out due to their ability to reclassify nearly 17\% of articles in the entire dataset (16.95\% uplift) from the undesired target class \textit{Evaluation = bad} to the desired target class \textit{Evaluation = good}. This highlights the substantial potential of these action rules to positively influence article evaluation outcomes on a large scale.

\subsubsection{CORD-19 Dataset - Citation Impact Enhancement} \label{sec:experiment4}

The second dataset analyzed in this study is CORD-19. The focus of this analysis is to identify action rules that could enhance the citation impact of articles. The following experiment is conducted:

\begin{table}[!ht]
    \centering
    \small
    \renewcommand{\arraystretch}{1.2} 
    \begin{tabular}{p{0.6cm}p{1.2cm}p{1.8cm}p{1.8cm}p{0.6cm}p{1cm}p{1.2cm}p{0.6cm}}
        \toprule
        \rotatebox{90}
        {\textbf{Rule}} & \rotatebox{90}{\textbf{Area}} & \rotatebox{90}{\textbf{Novelty}} & \rotatebox{90}{\textbf{Rigor}} & \rotatebox{90}{\textbf{Grammar}} & \rotatebox{90}{\textbf{Replicability}} & \rotatebox{90}{\textbf{Cited}} & \rotatebox{90}{\textbf{Uplift}} \\
        \midrule
        $r_{22}$ & educa. & & med $\rightarrow$ high & & 1 $\rightarrow$ 0 & 0 $\rightarrow$ 1 & 1.08\% \\
        $r_{23}$ & socia. & high & med $\rightarrow$ high & & 1 $\rightarrow$ 0 & 0 $\rightarrow$ 1 & 0.93\% \\
        $r_{24}$ & media. & high & med $\rightarrow$ high & 0 & 1 $\rightarrow$ 0 & 0 $\rightarrow$ 1 & 0.71\% \\
        $r_{25}$ & mathe. & high $\rightarrow$ med & med & & 1 $\rightarrow$ 0 & 0 $\rightarrow$ 1 & 0.65\% \\
        $r_{26}$ & langu. & high $\rightarrow$ med & med & & 1 $\rightarrow$ 0 & 0 $\rightarrow$ 1 & 0.62\% \\
        $r_{27}$ & langu. & high & med $\rightarrow$ high & & 1 $\rightarrow$ 0 & 0 $\rightarrow$ 1 & 0.57\% \\
        \bottomrule
    \end{tabular}
    \caption{Overview of the dominant action rules and their respective uplift percentages from the CORD-19 dataset.}
    \label{tab:action_rules_uplift_education_sociology}
\end{table}

The final experiment focuses on identifying action rules that can enhance the citation rate of articles. In this case, 16 action rules are discovered. The rule with the highest impact, as seen in Table \ref{tab:action_rules_uplift_education_sociology}, changes the classification of 1.08\% of articles in the dataset.

Among these 16 action rules, 6 dominant rules are identified, which are listed in Table \ref{tab:action_rules_uplift_education_sociology}. The effectiveness of the rules for improving citation rates is notably lower than for the M17+ dataset. This might be related to the higher quality of the M17+ dataset, where the target is based on expert assessment as contrasted with time-adjusted citation counts used in CORD-19.

\section{Discussion}\label{sec:discussion}
In this work, we evaluated the usability of the freely available Llama2 large language model for generating interpretable high-level features from text.

\textit{Feature generation} We used the base open-source Llama2 language model for the generation of text characteristics based on abstracts. These features are directly derived from the text, meaning, and context provided in the article abstracts. Many of these features would have to be assessed by a human evaluator, making the use of LLM a natural candidate for their extraction.  

\textit{Predictive performance compared to existing representations} First, we tested on two different datasets, which resulted in a marked increase in the accuracy over the baseline naive classifiers (50.2\% vs 59.8\% for CORD-19, 18\% vs 37\% for M17+).   This showed that LLM-generated features are informative, but how do they stand in comparison with standard text representation? 
In the case of predicting human evaluation (M17+), the model trained on LLM-generated features surpassed the classical TF-IDF-weighted BoW approach in terms of accuracy and was tied for the F1 score and MAE.

\textit{Fusion of LLM-features with other representations } 
The previously described experiments showed that LLM-generated features are informative, but would they enhance predictive accuracy beyond standard models? 
The results show that training machine-learning models on the LLM-generated features does increase predictive performance. Combining the LLM-generated features with the BoW approach for text classification further showed improvement over the use of plain BoW. Incorporating LLM-generated features into the state-of-the-art AutoGluon model in combination with text input improved its performance on both datasets, albeit providing non-explainable results.

\textit{Analysis of importance of individual LLM features} 
Interpretation of machine-learning models using SHAP indicated that \textit{research rigor} has the highest importance in predicting the citation rate (CORD-19) as well as in predicting human evaluation (M17+).
Interestingly, the effects of LLM-generated features were more pronounced in their relation to the human evaluation rather than citation rate. This is likely due to the human evaluation being performed by a jury of experts who take these characteristics of text into account, whereas previous bibliometric research \cite{tahamtan2016factors} indicates that the number of citations is also influenced by the bibliometric features such as journal impact factor or the number of authors, which are features that we did not include in our analysis.

The quality of LLM-generated features on the scientometric datasets could be further improved if full texts were used instead of abstracts, although the magnitude of this improvement is hard to predict. For example,  \cite{glenisson2005combining} observed only a slight improvement.

\textit{Accuracy-Interpretability tradeoff}
LLM-based feature generation is favourably positioned on the interpretability-accuracy spectrum. The interpretability is its core strength - the generated features have user-friendly and intuitively important features like \textit{Statistical analysis - None} (important for the M17+ dataset predicting research impact) or \emph{Presence of Racial Slurs - Yes} (important for the Hate speech detection dataset). These names can also be optionally described by the additional text, which could be very useful in the case of highly specialised features such as those present in the Food hazard dataset.
This is an advantage even compared to features commonly present in off-the-shelf datasets used for machine learning, which often lack these qualities \cite{gong2023survey,shome2022data}. 
In terms of accuracy, our results suggest that while classifier performance on LLM features may not reach the heights of deep learning methods that represent data in the latent space of embeddings, LLM features performed equally well or better against common interpretable baselines, such as TF-IDF.

It should be emphasized that LLM feature generation provides large opportunities for further improvements. For example, in our experiment focusing on the computational cost benchmark (Section~\ref{sec:llmonlocalgpus}), we used a chain of thought DeepSeek reasoning model. While it was out of the scope of this research to analyze its outputs in detail, an analysis of reasoning for excluding or including a specific feature could provide a viable alternative or complement to traditional approaches based on statistical feature selection, such as chi-square significance testing of forward feature selection. While not necessarily providing better results, the textual reasons will be better understandable for most users than numerical test results.

\textit{Rule learning over LLM-based features} In our previous research \cite{Beranova2022}, which involved rule mining over the BoW text representation of a subset of CORD-19 dataset, we struggled with the problem of highly specialized rules referring to individual rule conditions. This was partly addressed by rule clustering, which grouped similar rules together, but this had considerable limitations. The use of more general LLM-based features is a leap forward, as it is possible to define features at an arbitrary level of generality. For example, we had LLM generate a binary feature indicating whether the article abstract refers to any experiments. In \cite{Beranova2022}, we could infer the presence of experiments only from specific words, such as "mouse" that would suggest that experiments using murine models on animals were conducted.
Despite the presence of better features, the rule-based experiments in the present article exhibited some redundancy, but not on the level of rules. In the results, we often observed similar rules, which were often different only in one or two conditions and covered the same or highly similar set of articles. 
To counter this problem, the concept of dominant action rules was implemented, which removed rules representing mere extensions of an existing rule without any improvement in uplift. 

\section{Related work} \label{sec:relatedwork}
Despite the large number of works covering LLMs, only the arXiv preprints of \citet{zhang2024dynamic} and \citet{zhou2024llm} focus on LLM-based feature generation. Both of these preprints were made available shortly before this article was submitted.

\textit{Comparison to other LLM-based feature generation methods}

In \citet{zhang2024dynamic}, the authors describe a workflow, where LLM is used to generate agents for feature generation. This work is different from ours in several ways. First, it focuses on tabular datasets. Second, the evaluation is performed through predictive performance and no assessment of the explainability of generated features is included. Our approach is different, as it was specifically designed for text classification, and we also provide a detailed study of feature importance scores. We also provide results on the effect of combining LLM-based features with other feature representations.  

\citet{zhou2024llm} use LLMs to extract features for dialogue constructiveness assessment. The authors prompt an LLM to assess written dialogues or their parts. Features generated in this way include information content, dialogue tactics, quality of arguments or style. As a machine learning model, they use logistic and ridge regression.  This work is similar to ours in that the authors also use hand-defined feature names. In their case of discourse analysis, these are features such as formality or sentiment. In our case, relating to research articles, these are features such as rigor or novelty. The results of \citet{zhou2024llm} support our conclusions that LLM-based interpretable feature generation is feasible. 

In terms of methodology, our work uses an open-weight LLM (Llama 2), while \citet{zhang2024dynamic} use GPT 3.5 Turbo and \citet{zhou2024llm} use GPT 4, both closed models available through the OpenAI platform. Although \citet{zhou2024llm} notes that according to their informal experiments, Llama 3 results in poor annotation quality compared to GPT-4, we have used an open-source LLama 2 model with good success as demonstrated by the attained predictive performance measures. 

None of the earlier works performs a head-to-head comparison of the interpretability of LLM-based features and the bag of words (TF-IDF) features as we do. Crucially, neither of the two previous works aimed to generate LLM-based features for white-box model learning. The key contribution of our work is that we demonstrated the effectiveness of the use of LLM-generated features in rule learning. 

\textit{Comparison to alternative non-LLM feature generation methods}

The literature covers a large variety of feature extraction and generation methods. To scope the comparison, we focus on those that were used in conjunction with the datasets included in our study. 

Our prior research, \cite{Beranova2022,dvorackova2024explaining},  focused on building interpretable classifiers for the CORD-19 and M17+ datasets. These works utilized mainly rule-based methods and logistic regression; both would benefit from a less-dimensional and more interpretable feature set generated by LLMs.  

In related work utilizing the BANKING77 intent dataset, features are typically derived directly from the text via general-purpose NLP representations. For example, baseline models rely on token n‐gram frequencies or pretrained embeddings (fine-tuning a Transformer encoder like BERT or using a universal sentence encoder) to capture the semantic content of each query \citep{loukas2023breaking}.  \citet{he2022space} introduces a tree-structured contrastive learning framework and adaptive pre-training, which exploit the task-specific structure but demand significant pre-processing and tuning.
 
Hate Speech classifiers have historically leaned on explicitly crafted features: researchers often combine bag-of-words or TF–IDF weighted n-grams with manual lexicon indicators (e.g., presence of slurs or profanity) and other linguistic signals to detect hateful content \citep{almatarneh2019citius}. Deep neural approaches in this domain likewise incorporate word or sentence embeddings, but still benefit from such lexical features for improved recall of abusive language \citep{almatarneh2019citius}. \citet{markov2023holistic} applies key token probing and synthetic data augmentation tailored to the domain of toxic language, which improves performance but requires manual intervention and risks performance degradation due to noisy data.   
 
In contrast to most aforementioned approaches, the LLM-driven method described in this article generates interpretable features through prompt-based interaction without dataset-specific engineering.

\section{Conclusion} \label{sec:conclusion}
We have demonstrated a novel and effective approach using LLMs to extract low-dimensional, interpretable features from text, overcoming key limitations of traditional representations for interpretable machine learning. Our findings across diverse datasets confirm that these features offer competitive predictive performance compared to standard baselines like TF-IDF and even SciBERT embeddings, while critically retaining semantic meaning. This was particularly evident in the successful generation of understandable and actionable rules, demonstrating the practical utility for tasks demanding transparency. Both fully automated and user-specified feature generation workflows proved viable, offering flexibility for different use cases. This LLM-driven feature engineering paradigm opens promising avenues for applying a wider range of interpretable ML techniques directly to text.

\section*{Declarations}
\subsection*{Availability of code and data}
We used the implementation of machine learning classifiers from \texttt{scikit-learn} package \citep{scikit-learn}.

Code and data are available at \url{https://github.com/vojtech-balek/llm-features}. 

\subsection*{Competing interests}
The authors have no potential conflicts of interest to disclose.

\subsection*{Funding} 
The authors are grateful for the long-term support of research activities to the Faculty of Informatics and Statistics, Prague University of Economics and Business.

\subsection*{Author contributions}
V.B. data processing and analysis, L.S. action rule analysis and feature discovery pipeline, V.B., T.K., and L.S. writing, T.K., L.S., and V.B. methodology, V.S. scientometric expertise, M17+ dataset feature preparation, T.K. conceptualization, approval of the final manuscript, all authors.

The authors would like to thank Ngoc Bao Cap for help with preparing the M17+ dataset.

\subsection*{Declaration of generative AI and AI-assisted technologies in the writing process.}
During the preparation of this work, the authors used Grammarly, Gemini, and ChatGPT in order to improve the readability and language of the manuscript. After using this tool/service, the authors reviewed and edited the content as needed.

\begin{appendices}

\section{LLM prompts}\label{secA1}





\begin{figure}[H]
  \centering
  \begin{lstlisting}[
    basicstyle=\ttfamily\fontsize{6.9pt}{7.5pt}\selectfont,
    columns=flexible,
    breaklines=true,
    breakatwhitespace=true,
    frame=single,
    xleftmargin=10pt,
    linewidth=\textwidth]
{
  "system_message": "IMPORTANT: Return only a valid JSON object with no explanations, text, or markdown!!! Do not include any commentary or introductory text!!!",
  "input_metadata": {  
    "dataset_name": "$name",  
    "description": "$description",  
    "target": "$target",  
    "examples": "$examples"  
  },  
  "task": {  
    "steps": [  
      "Analyze the provided metadata and examples to determine the domain and context of the dataset.",  
      "Identify the key characteristics of the dataset relevant to predicting the target variable.",  
      "List potential high-level categorical and numerical features based on domain knowledge inferred from the dataset description.",  
      "Extract additional potential features from dataset examples using syntactic and semantic patterns, ensuring at least 20 distinct features are generated.",  
      "If the text implies certain values that match the target, these values may also be extracted as features. In cases where the target has multiple values, each value can be independently derived from the text as a feature if it is contextually appropriate.",  
      "For text-based datasets, identify key phrases, structural components, and linguistic patterns that are relevant.",  
      "For numerical datasets, identify aggregation patterns, distributional characteristics, and possible transformations.",  
      "Group related features into meaningful categories where applicable.",  
      "If a feature has more than 15 unique categories, group less frequent categories into an 'Other' class.",  
      "For each identified feature, provide a clear name, description, a complete list of possible values, and a specific LLM extraction query."  
    ],  
    "constraints": [  
      "Ensure features are distinct and non-redundant.",  
      "Note that the target variable is not explicitly present in the input text.",  
      "Prioritize domain-specific insights over generic ones.",  
      "Ensure output is a structured, valid JSON format.",  
      "For categorical variables, list possible values with domain justification.",  
      "For numeric variables, provide possible transformations (e.g., log, mean differences).",  
      "The extraction queries must be specific and detailed to ensure high-quality feature generation.",  
      "Tailor extraction queries to the domain context of the dataset.",  
      "Generate a diverse set of features to maximize potential predictive power."  
    ]  
  },  
  "output_format": {  
    "type": "json",  
    "structure": {  
      "features": [  
        {  
          "feature_name": "<Name of the categorical or numerical feature>",  
          "description": "<Short description of what the feature represents and how it relates to the dataset's context>",  
          "possible_values": ["<Value 1>", "<Value 2>", "...", "<Value n>"],  
          "extraction_query": "Identify the '<feature_name>' based on the provided context. Options: '<Value 1>', '<Value 2>', ..., '<Value n>'."
        }
      ]
    }
  }
}
  \end{lstlisting}
  \caption{Full prompt for LLM-based feature discovery. Prompt can be parameterised by inserting specific values instead of placeholders, which are denoted with \$ symbol.}
  \label{fig:fullprompt} 
\end{figure}

\begin{figure}[H]
  \centering
  \begin{lstlisting}[
    basicstyle=\ttfamily\fontsize{5.5pt}{5.6pt}\selectfont,
    columns=flexible,
    breaklines=true,
    breakatwhitespace=true,
    frame=single,
    xleftmargin=10pt,
    linewidth=\textwidth]
{
  "input_text": "$text",
  "task": "Extract the following features as described below and return a valid JSON object.",
  "constraints": [
    "The output must be a valid JSON.",
    "All answers must be simple and correspond to categorical values only."
  ],
  "features": [
    {
      "feature_name": "product_name",
      "description": "The name of the product being recalled, which can indicate the type of food and potential hazard.",
      "extraction_query": "Identify the 'product_name' based on the provided context. Options: 'Dole Fresh Blueberries', 'XSell Mini Vegetable Samosas', ..., 'Shikar brand'."
    },
    {
      "feature_name": "hazard_type",
      "description": "The type of hazard associated with the recalled product, which can indicate the nature of the risk to consumers.",
      "extraction_query": "Identify the 'hazard_type' based on the provided context. Options: 'biological', 'allergens', 'chemical', 'foreign bodies', 'fraud', 'packaging defect', 'food additives and flavourings', 'other hazard'."
    },
    {
      "feature_name": "recall_reason",
      "description": "The specific reason for the recall, which provides insight into the nature of the defect or contamination.",
      "extraction_query": "Identify the 'recall_reason' based on the provided context. Options: 'Cyclospora contamination', 'undeclared milk, soya & wheat', ..., 'incorrect packaging'."
    },
    {
      "feature_name": "distribution_area",
      "description": "The geographical area where the recalled product was distributed, which can indicate the scope of the recall.",
      "extraction_query": "Identify the 'distribution_area' based on the provided context. Options: 'United States', 'Canada', ..., 'Regional'."
    },
    {
      "feature_name": "company_name",
      "description": "The name of the company responsible for the recalled product, which can indicate the manufacturer or distributor involved.",
      "extraction_query": "Identify the 'company_name' based on the provided context. Options: 'Dole Diversified North America, Inc.', 'Heron Foods', ..., 'Universal Clearance Company'."
    },
    {
      "feature_name": "product_category",
      "description": "The category of the recalled product, which can indicate the type of food or beverage involved.",
      "extraction_query": "Identify the 'product_category' based on the provided context. Options: 'Fruits', 'Vegetables', ..., 'Packaged Foods'."
    },
    {
      "feature_name": "recall_date",
      "description": "The date when the recall was announced, which can indicate the timeliness of the recall action.",
      "extraction_query": "Identify the 'recall_date' based on the provided context. Format: 'YYYY-MM-DD'."
    },
    {
      "feature_name": "batch_code",
      "description": "The batch code of the recalled product, which can help identify the specific production lot affected.",
      "extraction_query": "Identify the 'batch_code' based on the provided context. Provide the alphanumeric code."
    },
    {
      "feature_name": "best_before_date",
      "description": "The best before date of the recalled product, which can indicate the shelf life and potential risk period.",
      "extraction_query": "Identify the 'best_before_date' based on the provided context. Format: 'YYYY-MM-DD'."
    },
    {
      "feature_name": "consumer_advice",
      "description": "The advice given to consumers regarding the recalled product, which can indicate the recommended actions to mitigate risk.",
      "extraction_query": "Identify the 'consumer_advice' based on the provided context. Options: 'Do not consume', 'Return to store', 'Dispose of product', 'Contact company'."
    }
  ],
  "output_format": {
    "type": "json",
    "structure": {
      "features": [
        {
          "feature_name": "<Feature Name>",
          "answer": "<Extracted Answer>"
        }
      ]
    }
  }
}
  \end{lstlisting}
  \caption{Example of an automatically generated prompt used for LLM-based feature generation for a single instance from the \textit{Food Hazard} dataset. Prompt can be parameterised by inserting specific text instead of the placeholder \$text.}
  \label{fig:genprompt} 
\end{figure}

\end{appendices}



\bibliography{bibliography.bib}

\end{document}